\documentclass{rsos}

\usepackage{graphicx}
\usepackage{latexsym}
\usepackage{epsfig}
\usepackage{multirow}
\usepackage{subfigure}
\usepackage{comment}
\usepackage{color}
\usepackage[usenames,dvipsnames]{xcolor}
\usepackage{bm}
\usepackage{mathrsfs}
\usepackage{enumitem}
\usepackage{hyperref}
\usepackage[normalem]{ulem}

\usepackage[sort&compress,numbers]{natbib}
\newcommand{\code}{\url{http://hdl.handle.net/10283/1924}}

\DeclareMathOperator{\Tr}{Tr}

\DeclareMathOperator\erf{erf}

\allowdisplaybreaks

\begin{document}

\title{Fast methods for training Gaussian processes on large data sets}

\def\addCambridge{Institute of Astronomy, Madingley Road, Cambridge, CB3 0HA, UK}
\def\addBirmingham{School of Physics \& Astronomy, University of Birmingham, Birmingham, B15 2TT, UK}
\def\addEdinburgh{School of Mathematics, University of Edinburgh and Biomathematics and Statistics Scotland, James Clerk Maxwell Building, Peter Guthrie Tait Road, Edinburgh EH9 3FD, UK}

\author{C.~J.~Moore$^{1}$, A.~J.~K.~Chua$^{1}$, C.~P.~L.~Berry$^{2}$ and J.~R.~Gair$^{3}$}

\address{$^{1}$\addCambridge\\$^{2}$\addBirmingham\\$^{3}$\addEdinburgh} 

%%%% Subject entries to be placed here %%%%
\subject{Statistics}

%%%% Keyword entries to be placed here %%%%

\corres{Christopher J. Moore\\
\email{cjm96@ast.cam.ac.uk}}

\begin{abstract}
Gaussian process regression (GPR) is a non-parametric Bayesian technique for interpolating or fitting data. The main barrier to further uptake of this powerful tool rests in the computational costs associated with the matrices which arise when dealing with large data sets. Here, we derive some simple results which we have found useful for speeding up the learning stage in the GPR algorithm, and especially for performing Bayesian model comparison between different covariance functions. We apply our techniques to both synthetic and real data and quantify the speed-up relative to using nested sampling to numerically evaluate model evidences.
\end{abstract}

\begin{fmtext}
\section{Introduction}

A wide range of commonly occurring inference problems can be fruitfully tackled using Bayesian methods. A particular common inference problem is that of regression; determining the relationship of a control variable $x$ to an output variable $y$ given a set of measurements of $\{y_{i}\}$ at points $\{x_{i}\}$. The solution requires a model $y = f(x)$, which allows us to predict the value of $y$ at an untested value of $x$. From a Bayesian standpoint, this can be achieved using Gaussian processes (GPs): a GP is collection of random variables, of which any finite subset have a joint Gaussian probability distribution \cite{GPR}.

\end{fmtext}

\maketitle

Gaussian process regression (GPR) is a powerful mathematical technique for performing non-parametric regression in a Bayesian framework \cite{GPR,Mackay,barry1986,wahba1978improper,o1978curve}. The key assumption underpinning the method is that the observed data set being interpolated is a realisation of a GP with a particular covariance function. This assumption presents us with a challenge: how do we choose the covariance function which gives the best interpolant?

The process of choosing the covariance function is known as \emph{learning}, or \emph{training} of the GP. In this training process, it is necessary to compute the inverse of the covariance matrix (the matrix formed by evaluating the covariance function pairwise between all $n$ observed points). The time taken to evaluate the inverse of the covariance matrix scales as ${\mathcal{O}}(n^{3})$ \cite{GPR}, where $n$ is the number of points being interpolated; this has typically restricted the application of GPR to smaller problems ($n\lesssim10^5$), although work has been done on extending its applicability to larger data sets \cite{Smola01sparsegreedy,Quinonero-Candela:2005:UVS:1046920.1194909,RSSB:RSSB633,banerjee2012efficient}.

In this paper, we present two techniques that speed up the training stage of the GPR algorithm. The first aims to reduce the dimensionality of the problem, and hence speed up the learning of the hyperparameters for a single covariance function; this does not change the fact that the cost of this process is ${\mathcal{O}}(n^{3})$, instead it simply reduces the constant in this scaling. The second aims to enable fast Bayesian model comparison between different covariance functions while also incorporating the benefits of the first technique.

We consider maximising the hyperlikelihood: the conditional probability of the data given a particular set of hyperparameters used to specify the covariance function.\footnote{This is often referred to as the marginal likelihood. In order to avoid confusion with distributions over the model parameters, we prefer to consistently use the hyper- prefix to denote probability distributions connected to the inference of the hyperparameters.} We provide an expression for the Hessian matrix of the hyperlikelihood surface and show how this can be used as a valuable tool for comparing the performance of two different covariance functions. We also present modified expressions for the hyperlikelihood, its gradient and its Hessian matrix, which have all been analytically maximised and marginalised over a single scale hyperparameter. This analytic maximisation or marginalisation reduces the dimensionality of the subsequent optimisation problem and hence further speeds up the training and comparison of GPs.

These techniques are useful when attempting to rapidly fit large, irregularly sampled datasets with a variety of covariance function models. The authors have previously made use of these techniques in exploring the correlation structure of the differences between complicated waveform models in the field of gravitational-wave astronomy \cite{2016PhRvD..93f4001M,2014PhRvL.113y1101M}; this was done so that the effect of different models on the parameter infererences could be marginalised over. There, the behaviour of the data was largely unknown a priori and it was necessary to quantitatively compare a wide range of different covariance functions. Work in this area with larger datasets is ongoing.

In Sec.~\ref{sec:GPR}, we review the GPR method and discuss methods of efficiently determining a covariance function. In Sec.~2\ref{sec:Hessian}, we present our expression for the Hessian of the hyperlikelihood along with a discussion of how it can be used for model comparison, and in Sec.~2\ref{sec:max} we show how the training of the GP can be accelerated by analytically maximising or marginalising the hyperlikelihood over a single scale parameter. In Sec.~\ref{sec:results}, we apply these methods to both synthetic and real data sets, and compare the computational cost to that of a full numerical evaluation of the Bayesian model evidences. Finally, a brief discussion and concluding remarks are given in Sec.~\ref{sec:dis}.

\section{Gaussian process regression \& training}\label{sec:GPR}

The technique of GPR is a method for interpolating (or extrapolating) the data contained in a \emph{training set} ${\mathcal{D}}=\small\{ \boldsymbol{x},\boldsymbol{y} \small\}$. The vector $ \boldsymbol{x}=\{ x_{i}| i=1,2,\ldots ,n \}$ is called the input vector and the output vector is given by $y_{i}=f(x_{i})$ for some unknown function $f$. The method works by assuming that the data have been drawn from an underlying GP $f(x)\sim {\mathcal{GP}}(\mu(x),k(x,x'))$ with specified mean $\mu(x)$ (usually assumed to be zero) and covariance function $k(x,x')$.

There is freedom in specifying the covariance function; common choices, such as the squared exponential and Mat\'ern function, include a number $m$ of free hyperparameters $\boldsymbol{\theta}=\{ \theta_{i}| i=1,2,\ldots ,m\}$ that control the properties of the GP, i.e.\ $k(x,x')=k(x,x';\boldsymbol{\theta})$.

The predictive power of the method comes from computing the conditional probability of the function taking a given value at some new $(n+1)$-th input point $x_{*}$, given the observed values in ${\mathcal{D}}$ and the values of the hyperparameters $\boldsymbol{\theta}$. This predictive probability distribution $P(y(x_{*})|{\mathcal{D}},\boldsymbol{\theta})$ for the function at the new point is a Gaussian with mean $\overline{y}(x_{*})$ and variance $[\sigma_y(x_*)]^2$ \cite{GPR},
\begin{align}
\overline{y}(x_{*}) = \boldsymbol{k}_{*}^\mathrm{T} \, {\mathbf{K}}^{-1} \, \boldsymbol{y}\,, \quad & \left[\sigma_y (x_*)\right]^2 =k_{**}-\boldsymbol{k}_{*}^\mathrm{T} \, {\mathbf{K}}^{-1} \, \boldsymbol{k}_{*}\,, \label{eq:meanvar}\\
y(x_{*})|\{\mathcal{D},\boldsymbol{\theta}\} \sim {} & {\cal{N}}\left(\overline{y}\left(x_{*}\right),\sigma_y\left(x_*\right)\right)\,, \label{eq:posterior}
\end{align}
where we have defined the scalar, vector and matrix shorthand 
\begin{equation}
k_{**} \equiv \; k(x_{*},x_{*}), \quad [ \boldsymbol{k}_{*}] _{i} \equiv k(x_{*},x_{i}) \; , \quad [ {\mathbf{K}}] _{ij} \equiv k(x_{i},x_{j}) \,.
\end{equation}
Since the posterior distribution for \eqref{eq:posterior} relies upon the form of the covariance, GPR cannot be used to make definite predictions until we have fixed a method for dealing with the unknown hyperparameters $\boldsymbol{\theta}$.

Ideally, we would like to place a prior probability distribution on $\boldsymbol{\theta}$ and make predictions by evaluating the integral
\begin{equation}
P(y(x_{*})|{\mathcal{D}}) = \int \; \mathrm{d}\boldsymbol{\theta}\; P(y(x_{*})|{\mathcal{D}},\boldsymbol{\theta})P(\boldsymbol{\theta}|{\mathcal{D}}) =\int \; \mathrm{d}\boldsymbol{\theta}\; P(y(x_{*})|{\mathcal{D}},\boldsymbol{\theta})P(\boldsymbol{y}|\boldsymbol{x},\boldsymbol{\theta})P(\boldsymbol{\theta})\,,\label{eq:TEMP}
\end{equation}
where we have used Bayes' theorem to obtain the second equality. We have introduced the hyperlikelihood given by
\begin{equation}
\ln P(\boldsymbol{y}|\boldsymbol{x},\boldsymbol{\theta}) = -\frac{1}{2}\left[\boldsymbol{y}^\mathrm{T} \, {\mathbf{K}}^{-1} \, \boldsymbol{y} + \ln (\det{\mathbf{K}}) + n\ln (2\pi) \right]\,,\label{eq:ev}
\end{equation}
which encodes the probability that the observed (training) data were drawn from a GP with covariance function $k$. The integral \eqref{eq:TEMP} is almost always analytically intractable and prohibitively expensive to evaluate numerically. A common approximate approach is to use the most probable values of the hyperparameters $\hat{\boldsymbol{\theta}}$, which maximise $P(\boldsymbol{\theta}|{\mathcal{D}})$ \cite{MacKay1999,snelson2005sparse,quinonero2007approximation}.

Assuming the prior distribution is sufficiently flat (or uninformative) over the region of interest, this is equivalent to maximising the hyperlikelihood $P(\boldsymbol{y}|\boldsymbol{x},\boldsymbol{\theta})$. Under this approximation, the predictive distribution becomes
\begin{equation}
P(y(x_{*})|{\mathcal{D}}) \simeq P(y(x_{*})|{\mathcal{D}},\hat{\boldsymbol{\theta}}) \,, \label{eq:MAP}
\end{equation}
which is simply the Gaussian in \eqref{eq:posterior} with mean and variance evaluated at $\hat{\boldsymbol{\theta}}$. Implementing the above procedure requires numerically maximising the hyperlikelihood in \eqref{eq:ev}. This can be computationally expensive; in Sec.~2\ref{sec:Hessian} and Sec.~2\ref{sec:max}, we present methods for reducing the cost of maximising the hyperlikelihood.

\subsection{Using the gradient \& Hessian}\label{sec:Hessian}

The maximisation process may be accelerated if the gradient of the hyperlikelihood is known and a gradient-based algorithm, such as a conjugate gradient method \cite{snelson2005sparse,blum2013optimization}, can be used. The gradient of the logarithm of the hyperlikelihood is given by \cite{GPR}
\begin{equation}
\partial_{\boldsymbol{\theta}} \ln P(\boldsymbol{y}|\boldsymbol{x},\boldsymbol{\theta}) = \frac{1}{2}\boldsymbol{y}^\mathrm{T} \, {\mathbf{K}}^{-1} \cdot \partial_{\boldsymbol{\theta}}{\mathbf{K}} \cdot {\mathbf{K}}^{-1} \, \boldsymbol{y} - \frac{1}{2}\Tr({\mathbf{K}}^{-1}\cdot\partial_{\boldsymbol{\theta}}{\mathbf{K}}) \,. \label{eq:grad}
\end{equation}
This can be shown by differentiating \eqref{eq:ev} and making use of the standard results
\begin{equation}
\partial {\mathbf{K}}^{-1} = -{\mathbf{K}}^{-1}\cdot\partial {\mathbf{K}} \cdot{\mathbf{K}}^{-1}\,, \quad
\partial \left(\det {\mathbf{K}}\right) = \left(\det {\mathbf{K}}\right)\Tr({\mathbf{K}}^{-1}\cdot\partial{\mathbf{K}})\,.
\label{eq:invdet}
\end{equation}
The gradient in \eqref{eq:grad} is useful because the rate-determining step in computing the hyperlikelihood is computing the inverse matrix ${\mathbf K}^{-1}$ (usually achieved through a Cholesky decomposition in practice), which is an ${\mathcal{O}}(n^{3})$ operation. All other steps in \eqref{eq:ev} scale as ${\mathcal{O}}(n^{2})$ or less.\footnote{As described in a footnote in \cite{GPR}, the matrix--matrix products in \eqref{eq:grad} should not be evaluated directly, as this is an ${\mathcal{O}}(n^{3})$ operation. Rather, the first term should be evaluated in terms of matrix--vector products, and, in the second term, only the diagonal elements that contribute to the trace need to be calculated; these are both ${\mathcal{O}}(n^{2})$ operations.} Once the inverse has been calculated, the gradient in \eqref{eq:grad} may also be evaluated in ${\mathcal{O}}(n^{2})$; so in evaluating the hyperlikelihood for a large training set we can also get the gradient for negligible extra cost.

The procedure outlined above can be performed for multiple covariance functions, each yielding a different GP interpolant. It is therefore necessary to have a method of comparing the performance of different interpolants to decide which to use. One way to achieve this is to evaluate the (hyperprior-weighted) volume under the hyperlikelihood surface, the hyperevidence, and use this as a figure of merit for the performance. Evaluating this integral is prohibitive, so an approximation is to calculate the Hessian matrix of the $\ln P(\boldsymbol{y}|\boldsymbol{x},\boldsymbol{\theta})$ surface at the peak (the position and value of which have already been found) and to analytically integrate the resulting Gaussian. This procedure assumes flat (or slowly varying) hyperpriors in the vicinity of the peak, but this has already been assumed in going from \eqref{eq:TEMP} to \eqref{eq:MAP}. Differentiating the gradient in \eqref{eq:grad}, again making use of the results in \eqref{eq:invdet}, and evaluating the derivatives at the position of peak hyperlikelihood, $\boldsymbol{\theta}=\hat{\boldsymbol{\theta}}$, gives the Hessian, 
\begin{align} 
\left.\partial_{\boldsymbol{\theta}}\partial_{\boldsymbol{\theta}'} \ln P(\boldsymbol{y}|\boldsymbol{x},\boldsymbol{\theta})\right|_{\hat{\boldsymbol{\theta}}} = {} & -\frac{1}{2}\boldsymbol{y}^\mathrm{T}\left[
2{\mathbf{K}}^{-1}\cdot\partial_{\boldsymbol{\theta}}{\mathbf{K}}\cdot{\mathbf{K}}^{-1}\partial_{\boldsymbol{\theta}'}{\mathbf{K}}\cdot{\mathbf{K}}^{-1}-{\mathbf{K}}^{-1}\cdot\partial_{\boldsymbol{\theta}}\partial_{\boldsymbol{\theta}'}{\mathbf{K}}\cdot{\mathbf{K}}^{-1} \right] \boldsymbol{y} \nonumber\\*
 & +\frac{1}{2}\Tr\left({\mathbf{K}}^{-1}\cdot\partial_{\boldsymbol{\theta}}{\mathbf{K}}\cdot{\mathbf{K}}^{-1}\cdot\partial_{\boldsymbol{\theta}'}{\mathbf{K}}- {\mathbf{K}}^{-1}\cdot\partial_{\boldsymbol{\theta}}\partial_{\boldsymbol{\theta}'}{\mathbf{K}}\right)
 =-\mathbf{H}\;.
\label{eq:Hess}
\end{align}
This expression has the same advantages as the expression for the gradient; as the inverse of the covariance matrix has already been computed, the Hessian may be evaluated at negligible extra cost. %Defining ${\mathbf{H}}$ to be the negative of the Hessian found in \eqref{eq:Hess}, the
The hyperlikelihood surface may therefore be approximated by the Gaussian \cite{MacKay1996,MacKay1999}
\begin{equation}
\ln P(\boldsymbol{y}|\boldsymbol{x},\boldsymbol{\theta}) \approx \ln P(\boldsymbol{y}|\boldsymbol{x},\hat{\boldsymbol{\theta}}) -\frac{1}{2}\Delta\boldsymbol{\theta}^\mathrm{T}\cdot {\mathbf{H}} \cdot \Delta\boldsymbol{\theta} \,. \label{eq:Gaussapprox} 
\end{equation}
We seek the evidence, which is given by the following integral of the posterior, where we have specified a prior $\Pi (\boldsymbol{\theta})$ on the hyperparameters;
\begin{equation} \mathcal{Z}(\mathcal{D}) = \int\mathrm{d}\boldsymbol{\theta}\;\Pi(\boldsymbol{\theta})P(\boldsymbol{y}|\boldsymbol{x},\boldsymbol{\theta}) \, .  \end{equation}
Assuming the posterior is a sufficiently well peaked distribution, with peak at position $\boldsymbol{\theta}=\tilde{\boldsymbol{\theta}}$, the evidence may be written using the Laplace approximation \cite{Mackay} as
\begin{equation}
\mathcal{Z}(\mathcal{D}) \approx \Pi(\tilde{\boldsymbol{\theta}})P(\boldsymbol{y}|\boldsymbol{x},\tilde{\boldsymbol{\theta}})\sqrt{\frac{(2\pi)^{m}}{\det{\left(\boldsymbol{H}+\boldsymbol{H}_{\Pi}\right)}}}  \,. \label{eq:chrisaddedmehere}
\end{equation}
It is always possible to change the hyperparameterisation so that the prior is flat in which case the hyperposterior is proportional to the hyperlikelihood.\footnote{For example, if the original prior on the parameters $\boldsymbol{\theta}$ is $p(\boldsymbol{\theta})$, we can define $\theta'_i (\boldsymbol{\theta}) = \int_{-\infty}^{\theta_i} p(\theta_i | \theta_{i+1}, \cdots, \theta_{m}) {\rm d} \theta_i$ and then $p(\boldsymbol{\theta'})$ is a constant.} If such a hyperparameterisation has been chosen then $\Pi(\tilde{\boldsymbol{\theta}})=1/V$ (where $V$ is the hyperprior volume, or range of integration), $\boldsymbol{H}_{\Pi}=0$, and $\tilde{\boldsymbol{\theta}}=\hat{\boldsymbol{\theta}}$; therefore
\begin{equation}
\mathcal{Z}({\mathcal{D}}) \approx \frac{P(\boldsymbol{y}|\boldsymbol{x},\hat{\boldsymbol{\theta}})}{V} \sqrt{\frac{(2\pi)^{m}}{\det{\mathbf{H}}}} \,. \label{eq:figofmerit}
\end{equation} 
This expression is now invariant under further changes to the hyperparameter specification which preserve the property that the prior is constant. We use hyperparameterisations with flat hyperpriors as this choice uniquely specifies the approximation in Eq.~(\ref{eq:figofmerit}); although there remains the possibility that another hyperparameterisation exists in which the posterior is better approximated as a Gaussian. 

For two covariance functions, $k_{1}$ and $k_{2}$, the odds ratio may be defined as the ratio of the value of \eqref{eq:figofmerit} evaluated with $k_{1}$ to the value evaluated using $k_{2}$, and this may be used to discriminate among competing models. The hyperprior volume $V$ in \eqref{eq:figofmerit} acts as an Occam factor, penalising models with greater complexity \cite{Mackay}. Once suitable prior volumes have been fixed, the Hessian approximation to the hyperevidence is a computationally inexpensive means of comparing covariance functions.

The Hessian may also be used to provide error estimates for the hyperparameters; from \eqref{eq:Gaussapprox} it can be seen that the inverse of the Hessian is the covariance matrix of the maximum hyperlikelihood estimator of the hyperparameters.

\subsection{Partial analytic maximisation}\label{sec:max}

In general, covariance functions can be arbitrarily complicated, with large numbers of hyperparameters. Inevitably, simple covariance functions are the most prevalent in the literature. If there are a small number of hyperparameters, then even reducing the number of hyperparameters by one can have a great impact on the length of time taken to maximise the hyperlikelihood. In this section, we show how the hyperlikelihood for any covariance function, regardless of complexity, can be analytically maximised over an overall scale parameter, thereby reducing the number of remaining hyperparameters. We also generalise the expressions for the gradient and the Hessian found in Sec.~2\ref{sec:Hessian} to this case.

Consider the following transformation of the covariance, $k(\boldsymbol{x_{i}},\boldsymbol{x_{j}})\rightarrow \sigma_f^2k(\boldsymbol{x_{i}},\boldsymbol{x_{j}})$; substituting this into the expression for the hyperlikelihood gives,
\begin{equation}\label{eq:maxevuse}
\ln P(\boldsymbol{y}|\boldsymbol{x},\boldsymbol{\theta}) = -\frac{1}{2\sigma_f^2} \boldsymbol{y}^\mathrm{T} \, {\mathbf{K}}^{-1} \, \boldsymbol{y} - \frac{1}{2}\ln (\det{\mathbf{K}})-\frac{n}{2}\ln (2\pi\sigma_f^2) \,.
\end{equation}
This function always has a unique maximum with respect to variations in $\sigma_f^2$ at the position
\begin{equation}\label{eq:anmaxsigf}
\hat{\sigma}_f^2 = \frac{1}{n}\boldsymbol{y}^\mathrm{T} \, {\mathbf{K}}^{-1} \, \boldsymbol{y}\,;
\end{equation}
at this point the hyperlikelihood takes the value
\begin{equation}
\ln P_{\mathrm{max}}(\boldsymbol{y}|\boldsymbol{x},\boldsymbol{\vartheta}) = -\frac{n}{2}\ln\left(2\pi e \hat{\sigma}_f^2\right) - \frac{1}{2}\ln (\det{\mathbf{K}}) \,.\label{eq:evmax}
\end{equation}
Equation \eqref{eq:evmax} is to be considered as a function of the remaining $m-1$ hyperparameters $\boldsymbol{\vartheta}= \{\boldsymbol{\theta} \setminus \sigma_f\}$. The peak evidence may now be found more easily by numerically maximising $\ln P_{\mathrm{max}}$ in \eqref{eq:evmax} with respect to the remaining parameters $\boldsymbol{\vartheta}$. If a gradient-based algorithm is used, it is advantageous to have an analogous expression to \eqref{eq:grad} to give inexpensive derivatives. This can be found by differentiating \eqref{eq:evmax} with respect to $\boldsymbol{\vartheta}$, making use of the results in \eqref{eq:invdet},
\begin{equation}
\partial_{\boldsymbol{\vartheta}} \ln P_{\mathrm{max}}(\boldsymbol{y}|\boldsymbol{x},\boldsymbol{\vartheta}) = \frac{1}{2\hat{\sigma}_f^2}\boldsymbol{y}^\mathrm{T} \, {\mathbf{K}}^{-1} \cdot \partial_{\boldsymbol{\vartheta}}{\mathbf{K}} \cdot {\mathbf{K}}^{-1} \, \boldsymbol{y} - \frac{1}{2}\Tr\left({\mathbf{K}}^{-1} \cdot \partial_{\boldsymbol{\vartheta}}{\mathbf{K}}\right)\,.\label{eq:grad2}
\end{equation}
These are not the same as the derivatives in \eqref{eq:grad}.

As well as maximising, we can also consider marginalising over $\sigma_{f}$ \cite{MacKay1996}. As we are marginalising over a scale parameter we use the (improper) Jeffreys prior $P(\sigma_{f}) =c/\sigma_{f}$ \cite{JeffreysPrior1946}. The result is equal to the maximised form, up to a multiplicative constant,
\begin{equation}
P_{\mathrm{marg}}(\boldsymbol{y}|\boldsymbol{x},\boldsymbol{\theta})= \int_{0}^{\infty}\mathrm{d}\sigma_{f}\;\frac{c}{\sigma_{f}}P(\boldsymbol{y}|\boldsymbol{x},\boldsymbol{\theta}) = \frac{c}{2}\left(\frac{2e}{n}\right)^{n/2} \Gamma\left(\frac{n}{2}\right) P_{\mathrm{max}}(\boldsymbol{y}|\boldsymbol{x},\boldsymbol{\theta}) \,.
\end{equation}

As before, once the peak hyperlikelihood has been found, the Hessian at the peak position can aid in model comparison. In this case, the Hessian should be calculated using the second derivatives of $\ln P_{\mathrm{marg}}$. However, we may instead differentiate $\ln P_{\mathrm{max}}$, as this differs only by a constant which will cancel when using the Hessian to compare two models. Differentiating \eqref{eq:grad2} with respect to $\boldsymbol{\vartheta}'$,\footnote{Here, $\hat{\sigma}_f$ retains its maximum $\ln P$ value from \eqref{eq:anmaxsigf}, although the new maximum $\ln P_{\mathrm{marg}}$ value has actually now shifted to $(\hat{\sigma}'_f)^2 = n\hat{\sigma}_f^2/(n-1)$ due to the effect of the hyperprior. For large data sets ($n \gg 1$) the difference between the two is negligible (the hyperprior becomes uninformative as it is overwhelmed by the hyperlikelihood).} 
\begin{align}
\left.\partial_{\boldsymbol{\vartheta}}\partial_{\boldsymbol{\vartheta}'}\ln P_{\mathrm{marg}}\right|_{\hat{\boldsymbol{\vartheta}}} \propto {} & \frac{1}{2n\hat{\sigma}_{f}^{4}}\boldsymbol{y}^\mathrm{T} \, {\mathbf{K}}^{-1} \cdot \partial_{\boldsymbol{\vartheta}}{\mathbf{K}} \cdot {\mathbf{K}}^{-1} \, \boldsymbol{y} \times \boldsymbol{y}^\mathrm{T}{\mathbf{K}}^{-1} \cdot \partial_{\boldsymbol{\vartheta}'}{\mathbf{K}} \cdot {\mathbf{K}}^{-1} \, \boldsymbol{y} \nonumber \\*
 & - \frac{1}{2\hat{\sigma}_f^2}\boldsymbol{y}^\mathrm{T}  \left[2\mathbf{K}^{-1} \cdot \partial_{\boldsymbol{\vartheta}}\mathbf{K} \cdot \mathbf{K}^{-1} \cdot \partial_{\boldsymbol{\vartheta}'}\mathbf{K} \cdot \mathbf{K}^{-1} - \mathbf{K}^{-1} \cdot \partial_{\boldsymbol{\vartheta}}\partial_{\boldsymbol{\vartheta}'}\mathbf{K} \cdot \mathbf{K}^{-1}\right]  \boldsymbol{y} \nonumber\\*
& + \frac{1}{2}\Tr\left(\mathbf{K}^{-1} \cdot \partial_{\boldsymbol{\vartheta}}\mathbf{K} \cdot \mathbf{K}^{-1} \cdot \partial_{\boldsymbol{\vartheta}'}\mathbf{K} - \mathbf{K}^{-1} \cdot \partial_{\boldsymbol{\vartheta}}\partial_{\boldsymbol{\vartheta}'}\mathbf{K} \right)\;.\label{eq:Hess2}
\end{align}
Again, these are not the same as the derivatives in \eqref{eq:Hess}. These expressions for the gradient and the Hessian of the hyperlikelihood, maximised or marginalised over $\sigma_f^2$, share the same advantages as the analogous expressions in Sec.~\ref{sec:GPR}: they may be evaluated in ${\mathcal{O}}(n^{2})$ time once the hyperlikelihood itself has been evaluated in ${\mathcal{O}}(n^{3})$ time.

\section{Numerical results}\label{sec:results}

In order to perform model comparison calculations between competing covariance functions, we must first specify at least two different covariance functions. We choose the two functions in \eqref{eq:k1_onetime} and \eqref{eq:k2_twotime}, where $(t,t')\equiv(x,x')$. These functions are both based on the periodic covariance function proposed by \cite{Mackay}. The first function $k_{1}$ is the product of a single periodic component with timescale $T_{1}$ and a simple compact-support polynomial covariance function \cite{Wendland} to describe any non-periodic component of the data. The choice of a compact-support covariance function is especially useful when working with large datasets; this is precisely the situation where the techniques described above are also designed to be of maximum benefit. The second function $k_{2}$ includes an additional periodic component with timescale $T_{2}$. In order to avoid double-counting in $k_{2}$, we impose the constraint $T_{2}\geq T_{1}$. Both covariance functions also include an uncorrelated noise term; we define this in such a way that $\sigma_{f}$ remains an overall scale hyperparameter which can be maximised or marginalised over analytically as described in Sec.~3\ref{sec:max}.
\begin{align}
\label{eq:k1_onetime}
k_{1}(t,t') = {} & \sigma_{f}^{2} C\left(\frac{|t-t'|}{T_{0}}\right) \exp\left[ - \frac{2}{l_{1}^{2}}\sin^{2}\left(\frac{\pi(t-t')}{T_{1}}\right) \right]+\sigma_{f}^{2}\sigma_{n}^{2}\delta_{tt'} \;, \\
\label{eq:k2_twotime}
k_{2}(t,t') = {} & \sigma_{f}^{2} C\left(\frac{|t-t'|}{T_{0}}\right) \exp\left[ - \frac{2}{l_{1}^{2}}\sin^{2}\left(\frac{\pi(t-t')}{T_{1}}\right) -\frac{2}{l_{2}^{2}}\sin^{2}\left(\frac{\pi(t-t')}{T_{2}}\right) \right]+\sigma_{f}^{2}\sigma_{n}^{2}\delta_{tt'} \;, \\
C(\tau) = {} & \begin{cases}
\displaystyle (1-\tau)^{5}\frac{48\tau^{2}+15\tau+3}{3} & \tau<1\\
0 & \tau > 1
\end{cases}.
\end{align}
The covariance functions are completely specified by the hyperparameters $\sigma_{f}$ (overall scale), $T_{j}$ ($j=0,1,2$; timescales), and $l_{j}$ ($j=1,2$; smoothing parameters for the periodic components). The noise parameter $\sigma_{n}$ could also be taken to be a hyperparameter; instead, for simplicity, we here take $\sigma_{n}$ to be fixed. As $\sigma_{n}$ appears in $k$ multiplied by the overall scale, $\sigma_f$, fixing $\sigma_{n}$ is roughly equivalent to specifying a fixed fractional error.

We want to perform model comparison using the Laplace approximation outlined previously. This technique requires reparametrising the covariance function such that the hyperpriors are flat. For the timescale hyperparameters, which are dimensionful, we choose to use the scale-invariant Jeffreys prior, $P(T_{j}) \propto 1/T_{j}$. This prior is improper if the range of $T_j$ is $(0,\infty)$, so we restrict the range to $(\delta t,\Delta T)$, where $\delta t$ and $\Delta T$ are respectively the smallest and largest separations between the sampling points. If there was a timescale in the problem outside of this range, we would be unable to resolve it from the data. We now seek a transformation $\phi_{j}\equiv\phi_{j}(T_{j})$ to a new hyperparameter $\phi_{j}$ such that the prior is flat in this parameter, $P(\phi_{j})=\mathrm{const}$. The conservation of probability gives a differential equation relating the two
\begin{equation}
P(T_{j})\mathrm{d}T_{j} = P(\phi_{j})\mathrm{d}\phi_{j} \quad \Rightarrow \quad T_{j}=\exp(\phi_{j}/A_j)\,,\quad \left\{j=0,1,2\right\}\,,
\end{equation}
where the $A_j$s are constants which we can set equal to $1$. The range of these new hyperparameters is $\phi_{j} \in(\ln(\delta t),\ln(\Delta T))$ and $P(\phi_j) = 1/\ln(\Delta T/\delta t)$.

For the smoothness parameters $l_{j}$ we choose to use log-normal priors, $P(l_{j})=\exp\left[-(\mu-\log l_{j})^{2}/(2\sigma_l^{2})\right]/\sqrt{2\pi\sigma_l^{2}}$, with mean $\mu=1$ and variance $\sigma_l^{2}=4$. As before, we seek a transformation to some new hyperparameters $\xi_{j}$ in which the prior is flat. The desired transformation is given by
\begin{equation}
l_{j}= \exp\left[\mu+\sqrt{2}\sigma_l\erf^{-1}(2\xi_{j})\right]\,,\quad \left\{j=1,2\right\}\,,
\end{equation}
where $\xi_{j}\in(-0.5,0.5)$.

\subsection{Synthetic data}\label{sec:synthetic}

\begin{figure}[h]
 \centering
 \includegraphics[trim=2.2cm 0.3cm 1.8cm 0.cm, width=0.7\textwidth]{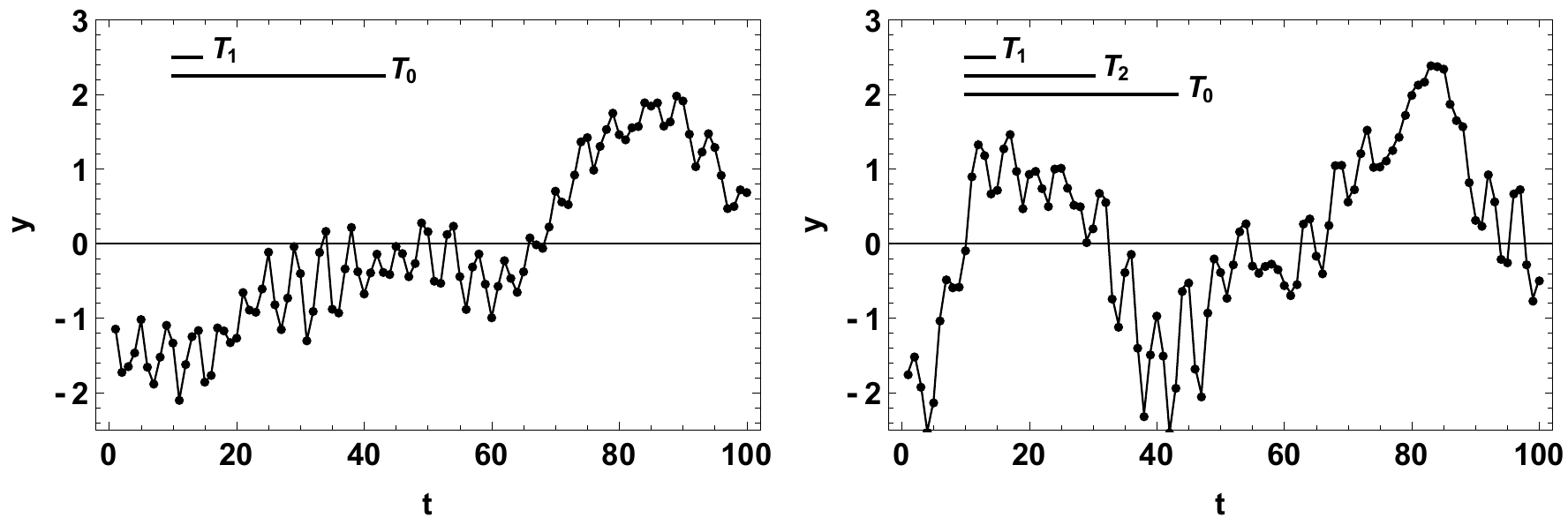}
 \caption{Realisations of the GPs $k_{1}(t,t')$ and $k_{2}(t,t')$ from \eqref{eq:k1_onetime} and \eqref{eq:k2_twotime} for values of $t=1,2,3\ldots,100$ are shown in the left and right hand panels respectively. The horizontal black lines indicate the lengthscales associated with the different terms in the covariance functions. The hyperparameters for $k_{1}$ where chosen to be $\sigma_{f}=1$, $\phi_{0}=3.5$, $\phi_{1}=1.5$, and $x_{1}=0$. The hyperparameters for $k_{2}$ where chosen to be the same as for $k_{1}$ and $\phi_{2}=3$ and $x_{2}=0$. In both cases the noise was fixed to $\sigma_{n}=10^{-2}$
 \label{fig:figsynthetic}}
\end{figure}

Shown in Fig.~\ref{fig:figsynthetic} are realisations of GPs with covariance functions $k_{1}$ and $k_{2}$.\footnote{The code, optimised for use on a GPU, and the synthetic used to produce these numerical results is available at \code.} In order to perform test model comparison calculations, a realisation of the $k_{2}$ GP with $n$ points was drawn and analysed using both the $k_{1}$ and $k_{2}$ covariance functions. For each covariance the peak hyperlikelihood was found by numerically maximising \eqref{eq:maxevuse} using a conjugate gradient method, making use of the gradient in \eqref{eq:grad2}. The hyperevidence was estimated using \eqref{eq:figofmerit} and the expression for the Hessian in \eqref{eq:Hess2}; the results are summarised in Tab.~\ref{tab:one}. To verify the accuracy of this estimate, the hyperevidence was also integrated numerically using {\sc MultiNest}, \cite{2009MNRAS.398.1601F,2008MNRAS.384..449F,2013arXiv1306.2144F} which implements a nested sampling algorithm \cite{skilling2006}. This was repeated for three different values of $n$ (in the case $n=100$ the synthetic data is plotted in the right-hand panel of Fig.~\ref{fig:figsynthetic}), and the results are also summarised in Tab.~\ref{tab:one}. 

\begin{figure}[h]
 \centering
 \includegraphics[trim=0cm 0cm 0cm 0cm, width=0.95\textwidth]{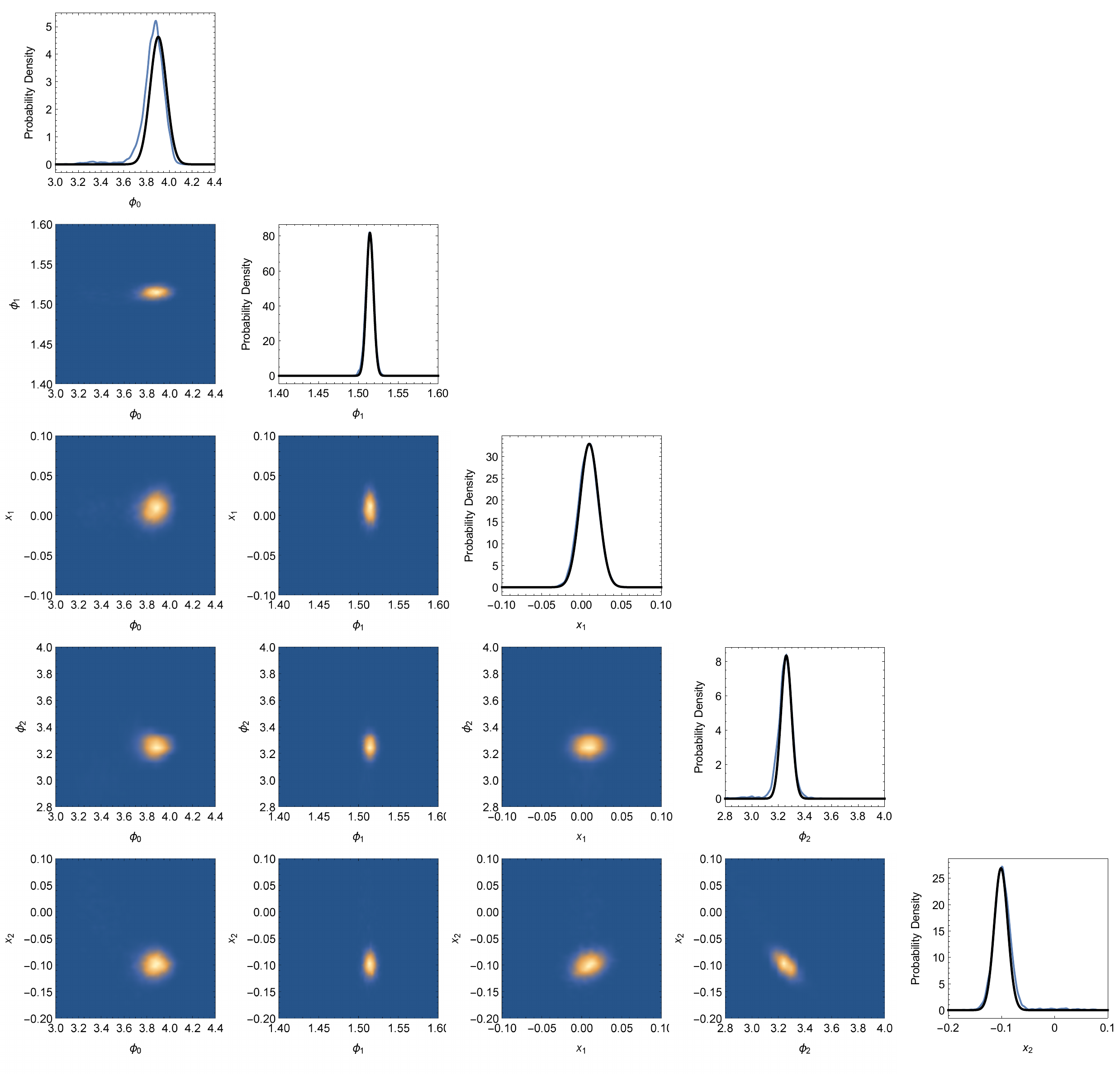}
 \caption{The one and two dimensional marginalised posterior distributions on the hyperparameters of the $k_{2}$ covariance function obtained from the largest ($n=300$) synthetic data set. The posterior is well approximated as a normal distribution. Shown in the black curves in the one dimensional marginalised posterior distributions along the diagonal are the normal approximations obtained by using the techniques described in Sec.~\ref{sec:GPR} to maximise and find the Hessian. Using the Hessian to approximate the integral of this distribution (the hyperevidence) leads to an error of $\sim 10\%$, see Tab.~\ref{tab:one}.\label{fig:figpost}}
\end{figure}

From Tab.~\ref{tab:one} it can be seen that as $n$ is increased, the Bayes factors increasingly favour the more complicated covariance function (and in this case the correct covariance function from which the data was drawn). In almost all cases, the Laplace approximation gives a value $\ln \mathcal{Z}_{\mathrm{est}}$ which is in agreement at better than $2\sigma$ with the numerically integrated value $\ln \mathcal{Z}_{\mathrm{num}}$. There is one exception which is highlighted in bold; this occurs for the most complicated covariance function (with the largest number of hyperparameters) and when the number of data points is smallest. In this situation, it would be expected that the posterior distribution on the hyperparameters may be highly multimodal and/or exhibit strong degeneracies (both of these expectations were confirmed by examining the posterior distribution on the hyperparameters returned by \textsc{MultiNest}). This exceptional case serves to highlight situations in which the Laplace approximation should not be trusted. The \textsc{MultiNest} posteriors in all other cases were verified to be well approximated by a single Gaussian mode. Fig.~\ref{fig:figpost} shows the posterior distribution for the parameters of $k_{2}$ obtained from the largest ($n=300$) synthetic data set.

Our method of model comparison is proposed as a faster alternative to model comparison using numerically evaluated Bayes factors. Simply comparing the peak hyperlikelihood (marginal likelihood) values would also give a measure of the goodness-of-fit, but this tends to favour more complex models and incurs the risk of overfitting. More sophisticated methods of model selection exist in the literature (see \cite{o2009review,piironen2015comparison} and references within), e.g., the comparison of models based on estimated predictive criteria \cite{geisser1979predictive,watanabe2009algebraic,Gelman2013}, or the construction of a larger reference model and the subsequent selection of a simpler submodel with similar predictions \cite{lindley1968choice,san1984predictive}. A detailed numerical comparison to these methods is left for future investigation.

The $\ln \mathcal{Z}_{\mathrm{num}}$ values in Tab.~\ref{tab:one}, evaluated using {\sc MultiNest}, required between 20,000 and 50,000 likelihood evaluations. The maximisation routines typically took fewer than $100$ likelihood evaluations to find the peak, and then one additional evaluation to calculate the Hessian and hence $\ln \mathcal{Z}_{\mathrm{est}}$. In order to guard against the possibility of the maximisation routines becoming trapped in local maxima, as opposed to the global maximum, the algorithm was run multiple times from randomly selected starting positions. The typical number of runs required to find the global maximum was $\sim10$. After these duplicate runs are accounted for, the speed-up factor in calculating $\ln \mathcal{Z}_{\mathrm{est}}$ compared to $\ln \mathcal{Z}_{\mathrm{num}}$ was between $20$ and $50$ in all cases.

\begin{table}
\begin{center}
{\scriptsize
\begin{tabular}{ c | c c | c c | c c }
  $n$ & $\ln \mathcal{Z}^{k_{1}}_{\mathrm{est}}$ & $\ln \mathcal{Z}^{k_{1}}_\mathrm{num}$ & $\ln \mathcal{Z}^{k_{2}}_{\mathrm{est}}$ & $\ln \mathcal{Z}^{k_{2}}_{\mathrm{num}}$ & $\ln {\mathcal{B}}_{\mathrm{est}}$ &  $\ln {\mathcal{B}}_{\mathrm{num}}$ \\
  \hline
  $30$   & $-17.77$ & $-17.87 \pm 0.08$ & $\mathbf{-18.82}$ & $-17.73 \pm 0.09$ & $-1.05$ & $0.14 \pm 0.12$ \\
  $100$  & $-20.17$ & $-20.17 \pm 0.10$ & $-19.22$ & $-19.22 \pm 0.11$ & $\hphantom{-}0.95$  & $0.95 \pm 0.15$ \\
  $300$  & $-49.94$ & $-50.12 \pm 0.11$ & $-40.21$ & $-40.36 \pm 0.13$ & $\hphantom{-}9.73$  & $9.76 \pm 0.17$ \\
\end{tabular}
}
\caption{A summary of the results of the analysis of synthetic data for three different-sized data sets. The first set of two columns is for a data set drawn from the $k_{2}$ covariance function and analysed with the $k_{1}$ covariance function. The first column is the estimated hyperevidence using the Laplace approximation where $\mathcal{Z}$ is as given in Eq.~\eqref{eq:figofmerit}, while the second is the numerically calculated hyperevidence. The second set of two columns shows results for the same data, but analysed with the $k_{2}$ covariance function. The final pair of columns shows the log Bayes factor, $\ln{\mathcal{B}}\equiv\ln \mathcal{Z}^{k_{2}}-\ln \mathcal{Z}^{k_{1}}$, calculated using the approximate and numerical values for the hyperevidence. \label{tab:one}}
\end{center}
\end{table}

\subsection{Tidal data from Woods Hole}

\begin{figure}[h]
 \centering
 \includegraphics[trim=2.2cm 0.3cm 1.8cm 0.cm, width=0.52\textwidth]{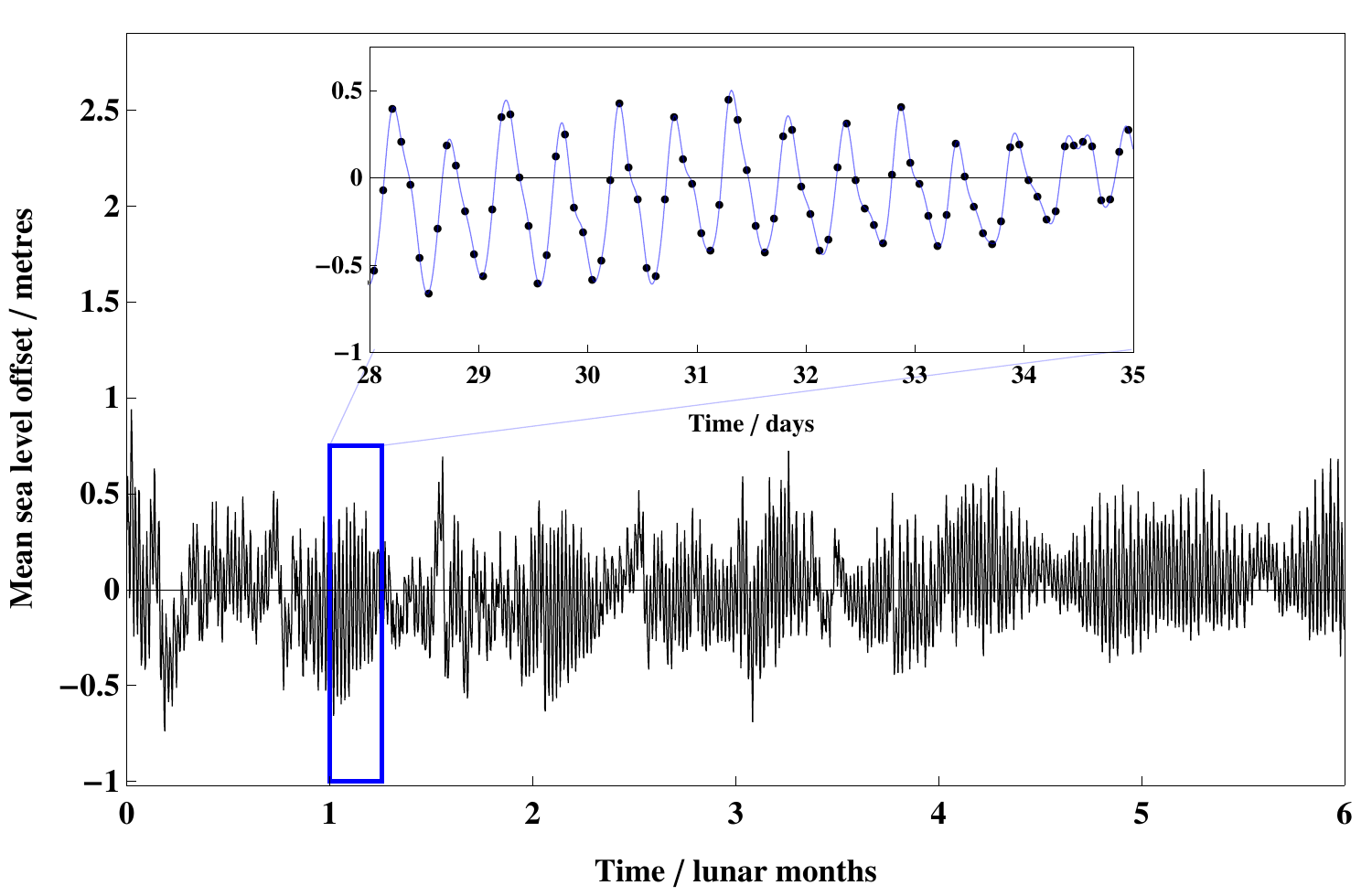}
 \caption{Shown in the main figure are six lunar months of tidal height data (black), from which the lunar tidal cycle can be discerned. Shown in the inset plot are several days of the tidal data (black points), from which the daily cycles can be clearly seen. Overlaid in the inset plot are both GP interpolants (blue), which are identical on this timescale.}
 \label{fig:fig}
\end{figure}

In order to illustrate the effectiveness of the techniques described above on real data, we consider several tidal data sets of different sizes from Woods Hole, MA \cite{tidaldata}.\footnote{Data from the National Oceanic and Atmospheric Administration, \url{http://tidesandcurrents.noaa.gov/waterlevels.html} (accessed August 2015).} We consider the mean sea level offset recorded between 3 January and 15 June 2014, six lunar months, sampled at two-hour intervals (giving $n=1968$ data points). This is plotted in Fig.~\ref{fig:fig}. We also consider a smaller subset of the data (the first lunar month), with $n=328$ data points.\footnote{This data set is regularly sampled in time, and therefore the covariance matrix will be a Toeplitz matrix. This structure could be exploited to accelerate the inversion of the covariance matrix; we choose not to use this here so that our code can be applied to irregularly sampled data in arbitrary dimensions.}

We interpolate the data using the two covariance functions in \eqref{eq:k1_onetime} and \eqref{eq:k2_twotime}; these functions are well suited to the data, as we expect the sea level to contain harmonics of the various timescales associated with the daily, monthly and yearly cycles of the tides. For simplicity, we fix $\sigma_{n}=10^{-2}$, which is the typical fractional error in the sea-level measurements. As in Sec.~3\ref{sec:synthetic} we reparametrise the covariance functions so that the $T_{j}$ have Jeffreys priors, and the smoothness parameters have log-normal priors. We use a conjugate gradient maximisation algorithm with \eqref{eq:grad2} and the Hessian in \eqref{eq:Hess2} to evaluate the volume in \eqref{eq:figofmerit} and perform model comparison between the two covariance functions. 

For the smaller dataset, we find the timescale $T_{1}=(12.8\pm0.2)~\mathrm{hours}$ with $k_1$, which corresponds to the two main tides per day. With $k_2$ we find the timescales $T_{1}=(12.44\pm0.07)~\mathrm{hours}$ and $T_{2}=(24.3\pm1.0)~\mathrm{hours}$; the second timescale corresponds to the height difference between the first and second tides of the day. The two-timescale model is highly favoured with a log Bayes factor of $57.8$.

For the larger dataset, we find $T_{1}=(12.80\pm0.11)~\mathrm{hours}$ with $k_1$, and $T_{1}=(12.40\pm0.03)~\mathrm{hours}$ and $T_{2}=(23.3\pm0.3)~\mathrm{hours}$ with $k_2$. In all cases, the (squared) errors are estimated using the diagonal components of the inverse Hessian; it can be seen that the timescales are more precisely measured for the larger dataset, as expected. The two-timescale model is even more conclusively favoured for the larger dataset, with a log Bayes factor of $538$. We also find a number of subsidiary hyperlikelihood peaks associated with other timescales in the data, but all subsidiary peaks are strongly suppressed relative to the global peak (by at least $\Delta \ln P$ of $\sim 100$) and so we expect our Bayes factor estimates to be robust.

Sea-level data is known to contain a large number of different frequencies, which necessitates the use of harmonic analysis in tidal modelling; the number of constituents included in tide prediction calculations has increased from tens \cite{lisitzin1974sea} to thousands \cite{casotto2004fully} over the past century. Clearly any $k_2$-like covariance function with $<10$ timescales is simplistic, but the construction of a more detailed tidal model is beyond the scope of this paper.~\footnote{We have conducted preliminary investigations of a three-timescale model. The hyperlikelihood surface for this covariance function is more structured and non-Gaussian than for $k_2$ and $k_1$. Estimates of the Bayes factors indicate that the inclusion of additional timescales is favoured, as expected based on the known large number of modes present.}

The number of evaluations of \eqref{eq:figofmerit} needed to obtain these results was comparable to the numbers for the synthetic data discussed in Sec.~3\ref{sec:synthetic}. However, each evaluation here was more expensive ($\sim 10~\mathrm{s}$) due to the size of the data set. Based on the speed-ups found in Sec.~3\ref{sec:synthetic}, it would be expected that \textsc{multinest} would take up to $\sim 1~\mathrm{week}$ to calculate the Bayes factor.

Shown in the inset plot in Fig.~\ref{fig:fig} are the two interpolants from $k_{1}$ and $k_{2}$ for the larger dataset, which both perform equally well on the timescale of one week. These interpolants, which are the result of the regression analysis, may be used to estimate the tidal height at a time where a measurement is not available. 

\section{Summary}\label{sec:dis}

We have described some simple ways in which the computationally expensive training stage of implementing GPR can be accelerated. The analytic maximisation of the hyperlikelihood over a single scale hyperparameter of the covariance function aids in speeding up the maximisation of the hyperlikelihood by reducing the dimensionality of the problem; the advantages of this will be most keenly felt in (common) problems where relatively simple covariance functions are used. Meanwhile, the analytic evaluation of the Hessian matrix, either in the manner of \eqref{eq:Hess} or \eqref{eq:Hess2}, aids in speeding up the process of model comparison between different types of covariance function. We have successfully demonstrated these techniques on a synthetic data set where the data was drawn from one of two covariance functions under consideration. In the case of the synthetic data the size of the data set was limited to less than 300 points, so that the results could be verified by using the {\sc MultiNest} algorithm to numerically sample and integrate the posteriors. We also demonstrated the techniques by applying them to a larger real data set of mean sea level measurements, where the full {\sc MultiNest} calculation would have taken too long to perform. It is to be hoped that these techniques will aid in the wider application of GP methods to larger data sets.

\paragraph{Data accessibility} The original source of the tidal data is the National Oceanic and Atmospheric Administration (\url{http://tidesandcurrents.noaa.gov/waterlevels.html}, accessed August 2015). The subset of data used here, as well as our synthetic data sets and our code is available at \code.
\paragraph{Funding statement} CJM and CPLB are supported by the STFC. AJKC's work is supported by the Cambridge Commonwealth, European and International Trust. JRG's work is supported by the Royal Society.
\paragraph{Author contributions} CJM devised the study, jointly performed the analysis and wrote the manuscript. AJKC jointly performed the analysis and wrote the manuscript. CPLB helped with the analysis, and the writing and editing of the manuscript. JRG helped devise the study and edited the manuscript. All authors gave final approval for the publication.
\paragraph{Conflict of interest} We have no competing interests.

\bibliographystyle{apsrev}
\bibliography{bibliography}

\begin{thebibliography}{32}
\expandafter\ifx\csname natexlab\endcsname\relax\def\natexlab#1{#1}\fi
\expandafter\ifx\csname bibnamefont\endcsname\relax
  \def\bibnamefont#1{#1}\fi
\expandafter\ifx\csname bibfnamefont\endcsname\relax
  \def\bibfnamefont#1{#1}\fi
\expandafter\ifx\csname citenamefont\endcsname\relax
  \def\citenamefont#1{#1}\fi
\expandafter\ifx\csname url\endcsname\relax
  \def\url#1{\texttt{#1}}\fi
\expandafter\ifx\csname urlprefix\endcsname\relax\def\urlprefix{URL }\fi
\providecommand{\bibinfo}[2]{#2}
\providecommand{\eprint}[2][]{\url{#2}}

\bibitem[{\citenamefont{Rasmussen and Williams}(2006)}]{GPR}
\bibinfo{author}{\bibfnamefont{C.~E.} \bibnamefont{Rasmussen}}
  \bibnamefont{and} \bibinfo{author}{\bibfnamefont{C.~K.~I.}
  \bibnamefont{Williams}}, \emph{\bibinfo{title}{Gaussian Processes for Machine
  Learning}} (\bibinfo{publisher}{MIT Press}, \bibinfo{address}{Cambridge, MA},
  \bibinfo{year}{2006}).

\bibitem[{\citenamefont{MacKay}(2003)}]{Mackay}
\bibinfo{author}{\bibfnamefont{D.~J.~C.} \bibnamefont{MacKay}},
  \emph{\bibinfo{title}{{Information Theory, Inference and Learning
  Algorithms}}} (\bibinfo{publisher}{Cambridge University Press},
  \bibinfo{address}{Cambridge}, \bibinfo{year}{2003}).

\bibitem[{\citenamefont{Barry}(1986)}]{barry1986}
\bibinfo{author}{\bibfnamefont{D.}~\bibnamefont{Barry}}, \bibinfo{journal}{The
  Annals of Statistics} \textbf{\bibinfo{volume}{14}}, \bibinfo{pages}{934}
  (\bibinfo{year}{1986}).

\bibitem[{\citenamefont{Wahba}(1978)}]{wahba1978improper}
\bibinfo{author}{\bibfnamefont{G.}~\bibnamefont{Wahba}},
  \bibinfo{journal}{Journal of the Royal Statistical Society. Series B
  (Methodological)} \textbf{\bibinfo{volume}{40}}, \bibinfo{pages}{364}
  (\bibinfo{year}{1978}).

\bibitem[{\citenamefont{O'Hagan and Kingman}(1978)}]{o1978curve}
\bibinfo{author}{\bibfnamefont{A.}~\bibnamefont{O'Hagan}} \bibnamefont{and}
  \bibinfo{author}{\bibfnamefont{J.~F.~C.} \bibnamefont{Kingman}},
  \bibinfo{journal}{Journal of the Royal Statistical Society. Series B
  (Methodological)} \textbf{\bibinfo{volume}{40}}, \bibinfo{pages}{1}
  (\bibinfo{year}{1978}).

\bibitem[{\citenamefont{Smola and Bartlett}(2001)}]{Smola01sparsegreedy}
\bibinfo{author}{\bibfnamefont{A.~J.} \bibnamefont{Smola}} \bibnamefont{and}
  \bibinfo{author}{\bibfnamefont{P.~L.} \bibnamefont{Bartlett}}, in
  \emph{\bibinfo{booktitle}{Advances in Neural Information Processing
  Systems}}, edited by \bibinfo{editor}{\bibfnamefont{T.~K.}
  \bibnamefont{Leen}},
  \bibinfo{editor}{\bibfnamefont{T.}~\bibnamefont{Dietterich}},
  \bibnamefont{and} \bibinfo{editor}{\bibfnamefont{V.}~\bibnamefont{Tresp}}
  (\bibinfo{publisher}{MIT Press}, \bibinfo{address}{Cambridge, MA},
  \bibinfo{year}{2001}), vol.~\bibinfo{volume}{13}, pp.
  \bibinfo{pages}{619--625}.

\bibitem[{\citenamefont{Qui{\~{n}}onero-Candela and
  Rasmussen}(2005)}]{Quinonero-Candela:2005:UVS:1046920.1194909}
\bibinfo{author}{\bibfnamefont{J.}~\bibnamefont{Qui{\~{n}}onero-Candela}}
  \bibnamefont{and} \bibinfo{author}{\bibfnamefont{C.~E.}
  \bibnamefont{Rasmussen}}, \bibinfo{journal}{Journal of Machine Learning
  Research} \textbf{\bibinfo{volume}{6}}, \bibinfo{pages}{1939}
  (\bibinfo{year}{2005}).

\bibitem[{\citenamefont{Cressie and Johannesson}(2008)}]{RSSB:RSSB633}
\bibinfo{author}{\bibfnamefont{N.}~\bibnamefont{Cressie}} \bibnamefont{and}
  \bibinfo{author}{\bibfnamefont{G.}~\bibnamefont{Johannesson}},
  \bibinfo{journal}{Journal of the Royal Statistical Society: Series B
  (Statistical Methodology)} \textbf{\bibinfo{volume}{70}},
  \bibinfo{pages}{209} (\bibinfo{year}{2008}).

\bibitem[{\citenamefont{Banerjee et~al.}(2013)\citenamefont{Banerjee, Dunson,
  and Tokdar}}]{banerjee2012efficient}
\bibinfo{author}{\bibfnamefont{A.}~\bibnamefont{Banerjee}},
  \bibinfo{author}{\bibfnamefont{D.~B.} \bibnamefont{Dunson}},
  \bibnamefont{and} \bibinfo{author}{\bibfnamefont{S.~T.}
  \bibnamefont{Tokdar}}, \bibinfo{journal}{Biometrika}
  \textbf{\bibinfo{volume}{100}}, \bibinfo{pages}{75} (\bibinfo{year}{2013}),
  \eprint{1106.5779}.

\bibitem[{\citenamefont{{Moore} et~al.}(2016)\citenamefont{{Moore}, {Berry},
  {Chua}, and {Gair}}}]{2016PhRvD..93f4001M}
\bibinfo{author}{\bibfnamefont{C.~J.} \bibnamefont{{Moore}}},
  \bibinfo{author}{\bibfnamefont{C.~P.~L.} \bibnamefont{{Berry}}},
  \bibinfo{author}{\bibfnamefont{A.~J.~K.} \bibnamefont{{Chua}}},
  \bibnamefont{and} \bibinfo{author}{\bibfnamefont{J.~R.}
  \bibnamefont{{Gair}}}, \bibinfo{journal}{Physical Review D}
  \textbf{\bibinfo{volume}{93}}, \bibinfo{eid}{064001} (\bibinfo{year}{2016}),
  \eprint{1509.04066}.

\bibitem[{\citenamefont{{Moore} and {Gair}}(2014)}]{2014PhRvL.113y1101M}
\bibinfo{author}{\bibfnamefont{C.~J.} \bibnamefont{{Moore}}} \bibnamefont{and}
  \bibinfo{author}{\bibfnamefont{J.~R.} \bibnamefont{{Gair}}},
  \bibinfo{journal}{Physical Review Letters} \textbf{\bibinfo{volume}{113}},
  \bibinfo{eid}{251101} (\bibinfo{year}{2014}), \eprint{1412.3657}.

\bibitem[{\citenamefont{MacKay}(1999)}]{MacKay1999}
\bibinfo{author}{\bibfnamefont{D.~J.~C.} \bibnamefont{MacKay}},
  \bibinfo{journal}{Neural Computation} \textbf{\bibinfo{volume}{11}},
  \bibinfo{pages}{1035} (\bibinfo{year}{1999}).

\bibitem[{\citenamefont{Snelson and Ghahramani}(2006)}]{snelson2005sparse}
\bibinfo{author}{\bibfnamefont{E.}~\bibnamefont{Snelson}} \bibnamefont{and}
  \bibinfo{author}{\bibfnamefont{Z.}~\bibnamefont{Ghahramani}}, in
  \emph{\bibinfo{booktitle}{Advances in Neural Information Processing
  Systems}}, edited by \bibinfo{editor}{\bibfnamefont{Y.}~\bibnamefont{Weiss}},
  \bibinfo{editor}{\bibfnamefont{B.}~\bibnamefont{Sch\"{o}lkopf}},
  \bibnamefont{and} \bibinfo{editor}{\bibfnamefont{J.~C.} \bibnamefont{Platt}}
  (\bibinfo{publisher}{MIT Press}, \bibinfo{address}{Cambridge, MA},
  \bibinfo{year}{2006}), vol.~\bibinfo{volume}{18}, pp.
  \bibinfo{pages}{1257--1264}.

\bibitem[{\citenamefont{Qui{\~{n}}onero-Candela
  et~al.}(2007)\citenamefont{Qui{\~{n}}onero-Candela, Rasmussen, and
  Williams}}]{quinonero2007approximation}
\bibinfo{author}{\bibfnamefont{J.}~\bibnamefont{Qui{\~{n}}onero-Candela}},
  \bibinfo{author}{\bibfnamefont{C.~E.} \bibnamefont{Rasmussen}},
  \bibnamefont{and} \bibinfo{author}{\bibfnamefont{C.~K.~I.}
  \bibnamefont{Williams}}, in \emph{\bibinfo{booktitle}{Large-scale Kernel
  Machines}}, edited by
  \bibinfo{editor}{\bibfnamefont{L.}~\bibnamefont{Bottou}},
  \bibinfo{editor}{\bibfnamefont{O.}~\bibnamefont{Chapelle}},
  \bibinfo{editor}{\bibfnamefont{D.}~\bibnamefont{DeCoste}}, \bibnamefont{and}
  \bibinfo{editor}{\bibfnamefont{J.}~\bibnamefont{Weston}}
  (\bibinfo{publisher}{MIT Press}, \bibinfo{address}{Cambridge, MA},
  \bibinfo{year}{2007}), chap.~\bibinfo{chapter}{9}, pp.
  \bibinfo{pages}{203--223}.

\bibitem[{\citenamefont{Blum and Riedmiller}(2013)}]{blum2013optimization}
\bibinfo{author}{\bibfnamefont{M.}~\bibnamefont{Blum}} \bibnamefont{and}
  \bibinfo{author}{\bibfnamefont{M.}~\bibnamefont{Riedmiller}}, in
  \emph{\bibinfo{booktitle}{European Symposium on Artificial Neural Networks,
  Computational Intelligence and Machine Learning}} (\bibinfo{year}{2013}).

\bibitem[{\citenamefont{MacKay}(1996)}]{MacKay1996}
\bibinfo{author}{\bibfnamefont{D.~J.~C.} \bibnamefont{MacKay}}, in
  \emph{\bibinfo{booktitle}{Maximum Entropy and Bayesian Methods}}, edited by
  \bibinfo{editor}{\bibfnamefont{G.~R.} \bibnamefont{Heidbreder}}
  (\bibinfo{publisher}{Springer Netherlands}, \bibinfo{address}{Dordrecht},
  \bibinfo{year}{1996}), vol.~\bibinfo{volume}{62} of
  \emph{\bibinfo{series}{Fundamental Theories of Physics}}, pp.
  \bibinfo{pages}{43--59}.

\bibitem[{\citenamefont{Jeffreys}(1946)}]{JeffreysPrior1946}
\bibinfo{author}{\bibfnamefont{H.}~\bibnamefont{Jeffreys}},
  \bibinfo{journal}{Proceedings of the Royal Society A: Mathematical, Physical
  and Engineering Sciences} \textbf{\bibinfo{volume}{186}},
  \bibinfo{pages}{453} (\bibinfo{year}{1946}).

\bibitem[{\citenamefont{Wendland}(2005)}]{Wendland}
\bibinfo{author}{\bibfnamefont{H.}~\bibnamefont{Wendland}},
  \emph{\bibinfo{title}{Scattered Data Approximation (Cambridge Monographs on
  Applied and Computational Mathematics)}} (\bibinfo{publisher}{Cambridge
  University Press}, \bibinfo{year}{2005}).

\bibitem[{\citenamefont{{Feroz} et~al.}(2009)\citenamefont{{Feroz}, {Hobson},
  and {Bridges}}}]{2009MNRAS.398.1601F}
\bibinfo{author}{\bibfnamefont{F.}~\bibnamefont{{Feroz}}},
  \bibinfo{author}{\bibfnamefont{M.~P.} \bibnamefont{{Hobson}}},
  \bibnamefont{and}
  \bibinfo{author}{\bibfnamefont{M.}~\bibnamefont{{Bridges}}},
  \bibinfo{journal}{Monthly Notices of the the Royal Astronomical Society}
  \textbf{\bibinfo{volume}{398}}, \bibinfo{pages}{1601} (\bibinfo{year}{2009}),
  \eprint{0809.3437}.

\bibitem[{\citenamefont{{Feroz} and {Hobson}}(2008)}]{2008MNRAS.384..449F}
\bibinfo{author}{\bibfnamefont{F.}~\bibnamefont{{Feroz}}} \bibnamefont{and}
  \bibinfo{author}{\bibfnamefont{M.~P.} \bibnamefont{{Hobson}}},
  \bibinfo{journal}{Monthly Notices of the Royal Astronomical Society}
  \textbf{\bibinfo{volume}{384}}, \bibinfo{pages}{449} (\bibinfo{year}{2008}),
  \eprint{0704.3704}.

\bibitem[{\citenamefont{{Feroz} et~al.}(2013)\citenamefont{{Feroz}, {Hobson},
  {Cameron}, and {Pettitt}}}]{2013arXiv1306.2144F}
\bibinfo{author}{\bibfnamefont{F.}~\bibnamefont{{Feroz}}},
  \bibinfo{author}{\bibfnamefont{M.~P.} \bibnamefont{{Hobson}}},
  \bibinfo{author}{\bibfnamefont{E.}~\bibnamefont{{Cameron}}},
  \bibnamefont{and} \bibinfo{author}{\bibfnamefont{A.~N.}
  \bibnamefont{{Pettitt}}}, \emph{\bibinfo{title}{{Importance Nested Sampling
  and the MultiNest Algorithm}}} (\bibinfo{year}{2013}), \eprint{1306.2144}.

\bibitem[{\citenamefont{Skilling}(2006)}]{skilling2006}
\bibinfo{author}{\bibfnamefont{J.}~\bibnamefont{Skilling}},
  \bibinfo{journal}{Bayesian Analysis} \textbf{\bibinfo{volume}{1}},
  \bibinfo{pages}{833} (\bibinfo{year}{2006}).

\bibitem[{\citenamefont{O'Hara and Sillanp{\"a}{\"a}}(2009)}]{o2009review}
\bibinfo{author}{\bibfnamefont{R.~B.} \bibnamefont{O'Hara}} \bibnamefont{and}
  \bibinfo{author}{\bibfnamefont{M.~J.} \bibnamefont{Sillanp{\"a}{\"a}}},
  \bibinfo{journal}{Bayesian Analysis} \textbf{\bibinfo{volume}{4}},
  \bibinfo{pages}{85} (\bibinfo{year}{2009}).

\bibitem[{\citenamefont{{Piironen} and
  {Vehtari}}(2015)}]{piironen2015comparison}
\bibinfo{author}{\bibfnamefont{J.}~\bibnamefont{{Piironen}}} \bibnamefont{and}
  \bibinfo{author}{\bibfnamefont{A.}~\bibnamefont{{Vehtari}}},
  \emph{\bibinfo{title}{{Comparison of Bayesian predictive methods for model
  selection}}} (\bibinfo{year}{2015}), \eprint{1503.08650}.

\bibitem[{\citenamefont{Geisser and Eddy}(1979)}]{geisser1979predictive}
\bibinfo{author}{\bibfnamefont{S.}~\bibnamefont{Geisser}} \bibnamefont{and}
  \bibinfo{author}{\bibfnamefont{W.~F.} \bibnamefont{Eddy}},
  \bibinfo{journal}{Journal of the American Statistical Association}
  \textbf{\bibinfo{volume}{74}}, \bibinfo{pages}{153} (\bibinfo{year}{1979}).

\bibitem[{\citenamefont{Watanabe}(2009)}]{watanabe2009algebraic}
\bibinfo{author}{\bibfnamefont{S.}~\bibnamefont{Watanabe}},
  \emph{\bibinfo{title}{Algebraic geometry and statistical learning theory}},
  Cambridge Monographs on Applied and Computational Mathematics
  (\bibinfo{publisher}{Cambridge University Press},
  \bibinfo{address}{Cambridge}, \bibinfo{year}{2009}).

\bibitem[{\citenamefont{{Gelman} et~al.}(2013)\citenamefont{{Gelman}, {Hwang},
  and {Vehtari}}}]{Gelman2013}
\bibinfo{author}{\bibfnamefont{A.}~\bibnamefont{{Gelman}}},
  \bibinfo{author}{\bibfnamefont{J.}~\bibnamefont{{Hwang}}}, \bibnamefont{and}
  \bibinfo{author}{\bibfnamefont{A.}~\bibnamefont{{Vehtari}}},
  \bibinfo{journal}{Statistics and Computing} \textbf{\bibinfo{volume}{24}},
  \bibinfo{pages}{997} (\bibinfo{year}{2013}), \eprint{1307.5928}.

\bibitem[{\citenamefont{Lindley}(1968)}]{lindley1968choice}
\bibinfo{author}{\bibfnamefont{D.~V.} \bibnamefont{Lindley}},
  \bibinfo{journal}{Journal of the Royal Statistical Society. Series B
  (Methodological)} \textbf{\bibinfo{volume}{30}}, \bibinfo{pages}{31}
  (\bibinfo{year}{1968}).

\bibitem[{\citenamefont{{San Martini} and
  Spezzaferri}(1984)}]{san1984predictive}
\bibinfo{author}{\bibfnamefont{A.}~\bibnamefont{{San Martini}}}
  \bibnamefont{and}
  \bibinfo{author}{\bibfnamefont{F.}~\bibnamefont{Spezzaferri}},
  \bibinfo{journal}{Journal of the Royal Statistical Society. Series B
  (Methodological)} \textbf{\bibinfo{volume}{46}}, \bibinfo{pages}{296}
  (\bibinfo{year}{1984}).

\bibitem[{\citenamefont{Scherer et~al.}(2001)\citenamefont{Scherer, Stoney,
  Mero, O'Hargan, Gibson, Hubbard, Weiss, Varmer, Via, Frilot
  et~al.}}]{tidaldata}
\bibinfo{author}{\bibfnamefont{W.}~\bibnamefont{Scherer}},
  \bibinfo{author}{\bibfnamefont{W.~M.} \bibnamefont{Stoney}},
  \bibinfo{author}{\bibfnamefont{T.~N.} \bibnamefont{Mero}},
  \bibinfo{author}{\bibfnamefont{M.}~\bibnamefont{O'Hargan}},
  \bibinfo{author}{\bibfnamefont{W.~M.} \bibnamefont{Gibson}},
  \bibinfo{author}{\bibfnamefont{J.~R.} \bibnamefont{Hubbard}},
  \bibinfo{author}{\bibfnamefont{M.~I.} \bibnamefont{Weiss}},
  \bibinfo{author}{\bibfnamefont{O.}~\bibnamefont{Varmer}},
  \bibinfo{author}{\bibfnamefont{B.}~\bibnamefont{Via}},
  \bibinfo{author}{\bibfnamefont{D.~M.} \bibnamefont{Frilot}},
  \bibnamefont{et~al.}, \bibinfo{type}{Tech. Rep.} \bibinfo{number}{NOAA
  Special Publication NOS CO-OPS 1}, \bibinfo{institution}{National Oceanic and
  Atmospheric Administration}, \bibinfo{address}{Silver Spring, Maryland}
  (\bibinfo{year}{2001}).

\bibitem[{\citenamefont{Lisitzin}(1974)}]{lisitzin1974sea}
\bibinfo{author}{\bibfnamefont{E.}~\bibnamefont{Lisitzin}},
  \emph{\bibinfo{title}{Sea-level changes}} (\bibinfo{publisher}{Elsevier},
  \bibinfo{address}{Oxford}, \bibinfo{year}{1974}).

\bibitem[{\citenamefont{{Casotto} and {Biscani}}(2004)}]{casotto2004fully}
\bibinfo{author}{\bibfnamefont{S.}~\bibnamefont{{Casotto}}} \bibnamefont{and}
  \bibinfo{author}{\bibfnamefont{F.}~\bibnamefont{{Biscani}}}, in
  \emph{\bibinfo{booktitle}{AAS/Division of Dynamical Astronomy Meeting \#35}}
  (\bibinfo{year}{2004}), vol.~\bibinfo{volume}{36} of
  \emph{\bibinfo{series}{Bulletin of the American Astronomical Society}}, p.
  \bibinfo{pages}{862}.

\end{thebibliography}

\end{document}